\documentclass[10pt,twocolumn]{article}

\usepackage[margin=0.75in,columnsep=0.2in,includeheadfoot,heightrounded]{geometry}
\usepackage{amsmath,amssymb}
\usepackage{graphicx}
\usepackage{booktabs}
\usepackage{microtype}
\usepackage{xcolor}
\usepackage{enumitem}
\usepackage{float}
\usepackage[hidelinks,bookmarks=false]{hyperref}
\usepackage{fancyhdr}
\usepackage{tcolorbox}
\usepackage{fancyvrb}
\usepackage{tabularx}

\definecolor{headerblue}{HTML}{000000}

\fancypagestyle{firstpage}{%
  \fancyhf{}
  
  \fancyhead[L]{%
    {\footnotesize\textcolor{headerblue}{KidsNanny: Multimodal Content Moderation}}%
  }
  \fancyhead[R]{%
    {\footnotesize\textcolor{headerblue}{March 2026}}%
  }

  \fancyfoot[C]{%
    \parbox{\textwidth}{\centering\scriptsize\textcopyright~2026 KidsNanny. Licensed under CC-BY-4.0.}%
  }
}

\nocite{freepiknsfwdetector,falconsainsfw,adamcoddvitnsfw,nudenet,shieldgemma2,helff2025llavaguard,euaiact2024,thorn2023csam,weprotect2023,ncmec2023report,dosovitskiy2021vit,vaswani2017attention,powers2011evaluation,lorenzo2023grooming,patchin2020sextortion,passvggdataset,qu2024unsafebench,qu2025modalitygap}

\setlist[itemize]{leftmargin=1.5em,noitemsep,topsep=0pt}
\setlist[enumerate]{leftmargin=1.5em,noitemsep,topsep=0pt}

\begin{document}
\thispagestyle{firstpage}

\twocolumn[{%
\noindent{\LARGE\bfseries KidsNanny: A Two-Stage Multimodal Content Moderation Pipeline Integrating Visual Classification, Object Detection, OCR, and Contextual Reasoning for Child Safety\par}

\bigskip

\noindent Viraj Panchal, Tanmay Talsaniya, Parag Patel, Meet Patel

\smallskip
\noindent{\scriptsize KidsNanny Research Team, Vartit Technology Inc.\enspace Correspondence: \href{mailto:research@kidsnanny.ca}{research@kidsnanny.ca}}

\bigskip\hrule\medskip

\noindent{\large\bfseries Abstract}\medskip

\noindent
We present KidsNanny, a two-stage multimodal content moderation architecture designed for child safety applications. Unlike existing approaches that rely on single-modality image classification or monolithic vision-language models (VLMs), the KidsNanny architecture separates content moderation into two complementary stages: Stage~1 combines a vision transformer (ViT) with an object detection module for real-time visual content screening, producing a classification decision and structured object detection labels; these outputs are then routed --- as text, not raw pixels --- to Stage~2, which applies OCR-based text extraction and passes the combined inputs to a text-based 7B language model for contextual confirmation of the safety verdict. We evaluate KidsNanny on the Sexual category of UnsafeBench comprising 1,054 images under two architecturally aligned evaluation regimes: (i)~a vision-only regime that isolates Stage~1 performance, and (ii)~a multimodal regime that evaluates the full Stage~1+2 pipeline. In the vision-only regime, KidsNanny Stage~1 achieves 80.27\% accuracy and 85.39\% F1-score at 11.7\,ms inference, outperforming four vision-only baselines. In the multimodal regime, the full pipeline achieves 81.40\% accuracy and 86.16\% F1-score at 120\,ms --- a modest +1.13 percentage point accuracy gain over Stage~1 alone, at ${\approx}$+108\,ms latency cost, while remaining approximately 9$\times$ faster than ShieldGemma-2 and 34$\times$ faster than LlavaGuard. To evaluate text-awareness capabilities, we filter two evaluation subsets from the UnsafeBench test set: a \emph{text+visual} subset (257 images) and a \emph{text-only} subset (44 images where safety depends primarily on embedded text). On text-only images, the KidsNanny model achieves 81.82\% accuracy, 100.00\% recall (25/25 positives; small sample), and 75.76\% precision. ShieldGemma-2 achieves 84\% recall but with only 60\% precision due to high false positives, at approximately 9$\times$ higher latency. These results demonstrate that dedicated OCR-based reasoning provides the best balance of recall, precision, and efficiency on text-embedded threats relevant to child safety.

\vspace{1em}
}]

\addcontentsline{toc}{section}{Abstract}

\section{Introduction}

The rapid growth of digital platforms accessible to children has created an urgent need for automated content moderation systems capable of detecting unsafe material in real time. Children encounter harmful content across multiple modalities --- explicit images, cyberbullying text overlaid on photographs, grooming language embedded in memes, and contextually inappropriate material that requires understanding the relationship between visual and textual elements to classify correctly.

Existing content moderation approaches fall into two broad categories. Single-modality image classifiers, such as NudeNet~\cite{nudenet}, FalconsAI~\cite{falconsainsfw}, and AdamCodd's ViT-based detector~\cite{adamcoddvitnsfw}, achieve high accuracy on visual content classification but cannot detect threats conveyed through text embedded in images. At the other end of the spectrum, large vision-language models (VLMs) such as LlavaGuard~\cite{helff2025llavaguard} and ShieldGemma-2~\cite{shieldgemma2} offer broader safety coverage and natural language reasoning but process raw image pixels through billion-parameter vision encoders, resulting in inference times of hundreds of milliseconds to seconds --- impractical for real-time moderation of high-volume content streams.

A critical gap exists between these approaches: to the best of our knowledge, no existing published academic system integrates dedicated image classification, object detection, OCR-based text extraction, and contextual reasoning in a unified pipeline. More fundamentally, no existing benchmark evaluates the ability of content moderation systems to detect threats conveyed through text embedded within images --- a modality particularly relevant to child safety, where unsafe content frequently appears as text overlaid on otherwise innocuous images.

We address these gaps with KidsNanny, a two-stage multimodal content moderation pipeline. This paper is a first-party technical report: the system is proprietary, evaluation was conducted by the development team, and results have not been independently verified. The contributions below should be read in that context. Our contributions are as follows:

\begin{enumerate}
  \item \textbf{Two-stage architecture.} We present a content moderation pipeline that separates fast visual screening (Stage~1: ViT + object detection, 11.7\,ms) from deeper multimodal analysis (Stage~2: OCR + text-based 7B LLM reasoning, 120\,ms total). By routing only extracted text and object labels to the language model --- rather than raw pixels --- Stage~2 achieves efficient inference without requiring a computationally expensive vision-language model.

  \item \textbf{Two-regime evaluation.} We evaluate KidsNanny under two architecturally aligned regimes --- vision-only and multimodal --- ensuring fair comparison against both lightweight classifiers and vision-language models. This design isolates the contribution of each pipeline stage.

  \item \textbf{Benchmark evaluation.} In the vision-only regime, KidsNanny Stage~1 outperforms four vision-only baselines on the UnsafeBench Sexual category. In the multimodal regime, KidsNanny (Stage~1+2) surpasses both ShieldGemma-2 and LlavaGuard at 9$\times$ and 34$\times$ lower inference latency, respectively.

  \item \textbf{Text-subset analysis.} We filter text-containing subsets from UnsafeBench and demonstrate that KidsNanny's OCR-based reasoning achieves 100\% recall on text-only threats (25/25 positives; n\,=\,44 images; interpret cautiously given small sample), compared to 84\% for ShieldGemma-2 and 56\% for LlavaGuard --- while maintaining superior precision (75.76\% vs.\ 60\%) and approximately 9$\times$ lower latency.

  \item \textbf{Benchmark gap identification.} We identify the absence of dedicated text-in-image safety benchmarks as a critical gap and propose directions for future multimodal safety evaluation.
\end{enumerate}

\section{Related Work}

\subsection{Image Safety Classification}

Binary NSFW classifiers have evolved from traditional CNN architectures to transformer-based approaches. Early work used ResNet and Inception architectures trained on datasets like Yahoo's Open NSFW dataset. More recently, ViT-based classifiers such as AdamCodd's vit-base-nsfw-detector~\cite{adamcoddvitnsfw} and FalconsAI's NSFW Image Detection~\cite{falconsainsfw} report high accuracy on curated NSFW benchmarks. NudeNet~\cite{nudenet} takes a different approach using object detection to identify specific body parts, providing localization alongside classification but limited to nudity detection. Freepik's NSFW Image Detector~\cite{freepiknsfwdetector} applies a ViT backbone for binary content filtering.

These models excel at their designed task but share a fundamental limitation: they operate exclusively on visual signals and cannot detect threats conveyed through text embedded within images or through contextual relationships requiring multimodal reasoning.

\subsection{Vision-Language Models for Content Safety}

The emergence of large VLMs has enabled more sophisticated safety classification. LlavaGuard~\cite{helff2025llavaguard} fine-tunes a 7B-parameter LLaVA-OneVision model to perform policy-customizable safety assessment across 10+ harm categories, achieving strong performance but at inference times exceeding 4 seconds per image. ShieldGemma-2~\cite{shieldgemma2} fine-tunes Gemma~3 (4B parameters) for image safety classification across sexually explicit, violence, and dangerous content categories, introducing a novel adversarial data generation pipeline.

Notably, ShieldGemma-2's authors explicitly scope out this challenge: ``it is beyond the scope of our detector for this specific challenge of evaluating unsafe content that emerges pragmatically from the interplay of different modalities co-existing within one image,'' noting that ``a visually benign image, for instance, might be rendered unsafe by the specific meaning of text embedded within the image itself''~\cite{shieldgemma2}. This acknowledged limitation in a leading model directly motivates our dedicated OCR and text reasoning approach. Our text-subset analysis (Section~\ref{sec:text_subset}) provides quantitative confirmation of this limitation.

\subsection{Online Child Safety and Grooming Detection}

The intersection of AI and child safety extends beyond image classification. Research on online grooming detection has employed machine learning methods to identify predatory language patterns in chat logs~\cite{lorenzo2023grooming}. Reports from the National Center for Missing \& Exploited Children~\cite{ncmec2023report}, Thorn~\cite{thorn2023csam}, and the WeProtect Global Alliance~\cite{weprotect2023} document the increasing scale and sophistication of online threats targeting children.

\subsection{Multimodal Content Moderation}

Commercial content moderation systems deployed at scale typically use cascading pipelines of specialized models, but these architectures are proprietary and undocumented. Recent work addresses the \emph{modality gap} in VLMs --- the observation that these models detect unsafe textual concepts more reliably than unsafe visual ones --- by fine-tuning VLMs with reinforcement learning (PPO) to improve visual unsafe concept recognition~\cite{qu2025modalitygap}. This training-time approach complements KidsNanny's inference-time architecture: rather than retraining a single VLM, KidsNanny routes images through a fast visual stage first and invokes a text-only LLM only when text cues are detected. A significant evaluation gap also exists: current image safety benchmarks such as UnsafeBench~\cite{qu2024unsafebench} contain images with embedded text, yet, to the best of our knowledge, no prior work has separately evaluated model performance on these text-containing images or measured how well systems leverage embedded text for safety classification. KidsNanny contributes to both gaps --- presenting an explicit two-stage multimodal architecture and providing what we believe to be the first quantitative evaluation that isolates performance on text-containing subsets of existing benchmarks.

\section{Methodology}
\label{sec:methodology}

\subsection{System Architecture}
\label{sec:architecture}

The KidsNanny Moderation Model is a multimodal moderation system that integrates object detection, OCR-based text classification, and a reasoning layer to evaluate unsafe content in images. The architecture employs Transformer-based image classification backbones~\cite{dosovitskiy2021vit} for image-level feature extraction and CNN-based real-time object detection. An attention-driven reasoning module~\cite{vaswani2017attention} fuses visual, object, and textual signals to generate explainable safety decisions.

KidsNanny processes input images through a two-stage pipeline:

\textbf{Stage~1 --- Visual Screening:} A ViT image classifier and object detection module analyze the image's visual content. The classifier outputs a safety probability; the detector identifies safety-relevant objects with bounding boxes. Stage~1 processes every input image and provides real-time screening.

\textbf{Stage~2 --- Multimodal Analysis:} An OCR engine extracts text embedded within the image. The extracted text and object labels from Stage~1 are passed to a 7B language model, which performs contextual reasoning and produces a final safety verdict with a natural language explanation. The language model receives only text inputs --- not raw image pixels --- avoiding the computational overhead of vision-language models.

\begin{figure*}[!t]
\centering
\includegraphics[width=0.52\textwidth,height=0.50\textheight,keepaspectratio]{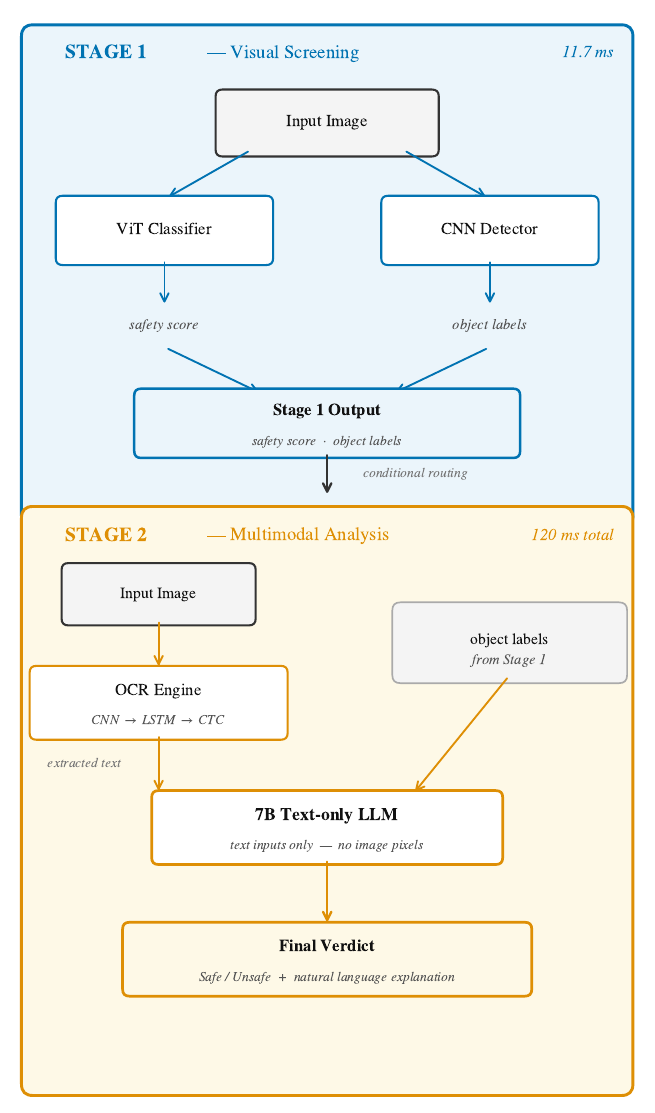}
\caption{KidsNanny two-stage architecture. Stage~1 performs visual screening using a ViT classifier and object detector (11.7\,ms). Stage~2 extracts text via OCR and passes extracted text + object labels (not raw pixels) to a 7B text-based language model for contextual reasoning (120\,ms total pipeline).}
\label{fig:architecture}
\end{figure*}

\textbf{Key architectural decision:} Stage~2's language model receives only text inputs --- object labels from Stage~1 and extracted text from the OCR engine --- rather than raw image pixels. This design choice enables 120\,ms full-pipeline inference versus 1,136--4,138\,ms for VLM-based approaches that process high-resolution image tokens through billion-parameter vision encoders.

\subsubsection{Architecture Specification}
\label{sec:arch_spec}

Tables~\ref{tab:stage1_components} and~\ref{tab:stage2_components} list the components of Stage~1 and Stage~2 respectively.

\begin{table*}[!t]
\centering
\caption{Stage~1 Components --- Visual Screening (11.7\,ms per image).}
\label{tab:stage1_components}
\small
\begin{tabular}{@{}llll@{}}
\toprule
\textbf{Component} & \textbf{Architecture} & \textbf{Input} & \textbf{Output} \\
\midrule
Image Classifier & ViT-based backbone & RGB image ($448\times448$) & Safety probability $\in [0,1]$ \\
Object Detector & CNN-based detector & RGB image ($512\times512$) & Bounding boxes + class labels + confidence \\
\bottomrule
\end{tabular}
\end{table*}

\begin{table*}[!t]
\centering
\caption{Stage~2 Components --- Multimodal Analysis (120\,ms total pipeline).}
\label{tab:stage2_components}
\footnotesize
\setlength\tabcolsep{4pt}%
\begin{tabular}{@{}p{4.0cm}p{3.2cm}p{4.6cm}p{4.8cm}@{}}
\toprule
\textbf{Component} & \textbf{Architecture} & \textbf{Input} & \textbf{Output} \\
\midrule
OCR Engine & Dedicated OCR module & RGB image (original resolution) & Extracted text with spatial \mbox{coordinates} \\[2pt]
Contextual Reasoning LLM & 7B LLM & Text: object labels + extracted text & Verdict (Safe/Unsafe) + explanation \\
\bottomrule
\end{tabular}
\end{table*}

Exact model variant names, parameter counts, and architectural hyperparameters are withheld as proprietary. Accordingly, the primary reproducible contribution of this paper is the two-regime evaluation methodology and text-subset benchmarking framework, both of which can be applied to any content moderation system evaluated on UnsafeBench. The model architecture as described --- a ViT-based visual classifier cascaded with an OCR-augmented text reasoner --- is intended to be representative of the design class rather than a fully specified replicable system.

\subsubsection{Two-Stage Inference Strategy}
\label{sec:inference_strategy}

Stage~1 operates on every input image at 11.7\,ms per image. Stage~2 is invoked conditionally --- only when Stage~1 flags an image as potentially unsafe or ambiguous, or when the OCR module detects text within the image. Images classified as clearly safe by Stage~1 and containing no embedded text skip Stage~2 entirely, keeping average latency close to Stage~1 for the majority of content.

\subsection{OCR-Based Text Safety}
\label{sec:ocr}
Many unsafe images contain minimal visual explicitness but include harmful text such as:
\begin{itemize}[leftmargin=1.5em]
  \item Sexual invitations or grooming language
  \item Coded or suggestive phrases
  \item Explicit words embedded in memes or screenshots
\end{itemize}

KidsNanny's OCR module extracts and analyzes this text, enabling detection of unsafe content that would bypass vision-only models. Research on online grooming linguistics~\cite{lorenzo2023grooming} confirms that textual cues are often the earliest indicators of predatory behavior, reinforcing the importance of text-aware moderation.

\subsection{Contextual Reasoning Module}
\label{sec:reasoning}
Beyond classification, KidsNanny integrates a context-aware reasoning layer that generates structured, human-interpretable reasoning for each moderation decision. This is especially important for nuanced threats such as sextortion, which has emerged as a rapidly growing risk among adolescents~\cite{patchin2020sextortion}. When unsafe or ambiguous content is detected, the model produces:
\begin{itemize}[leftmargin=1.5em]
  \item \textbf{Analysis:} A concise explanation of the visual, object, and textual signals that contributed to the decision.
  \item \textbf{Recommendation:} A clear moderation action (e.g., block, review, allow with warning) aligned with child-safety policy.
\end{itemize}

\subsection{Training Data}
\label{sec:training_data}

KidsNanny's pipeline components are trained on proprietary datasets curated for child safety content moderation. Training data sources and dataset statistics are withheld due to commercial confidentiality.

We confirm that: (1) no images from the UnsafeBench evaluation dataset or the PASS dataset were used during training, (2) no child sexual abuse material (CSAM) was used at any stage of development, and (3) all training data was collected and used in compliance with applicable data protection regulations.

\subsection{Model Selection Transparency}
\label{sec:model_selection}

During development, we evaluated multiple architectural configurations varying backbone architectures, object detection models, and inference strategies. The final model was selected based on performance on a held-out validation set disjoint from the UnsafeBench evaluation set. No hyperparameter tuning or architectural decisions were made using UnsafeBench test set performance.

\section{Evaluation}
\label{sec:evaluation}

\subsection{Benchmark}
\label{sec:benchmark}

We evaluate KidsNanny on the Sexual category of UnsafeBench~\cite{qu2024unsafebench}, a benchmarking framework for image safety classifiers covering 11 unsafe categories. The full dataset contains 10,146 images (4,048 unsafe and 6,098 safe). Our evaluation uses the Sexual category subset comprising 1,054 images (683 unsafe, 371 safe).

\subsubsection{PASS Dataset (False-Positive Validation)}

PASS~\cite{passvggdataset} is a large-scale dataset containing approximately 1.44 million images, intentionally excluding images containing humans. A random subset of 11,685 images was used as a clearly safe control set. All evaluated models achieve near-saturated performance (approximately 99\% specificity) on PASS, confirming calibration stability. Given this limited discriminative value, the primary comparative evaluation is based on UnsafeBench.

\subsection{Two-Regime Evaluation Design}
\label{sec:two_regime}

To ensure architecturally fair comparisons, we conduct two evaluation regimes on the same 1,054-image test set:

\textbf{Regime~1 --- Vision-Only.} All models operate purely on image pixels. For KidsNanny, OCR and contextual reasoning (Stage~2) are disabled; only Stage~1 (ViT classifier + object detector) produces the classification decision. Comparison models: NudeNet~\cite{nudenet}, FalconsAI~\cite{falconsainsfw}, Adam-ViT~\cite{adamcoddvitnsfw}, and Freepik~\cite{freepiknsfwdetector}.

\textbf{Regime~2 --- Multimodal.} All models have access to text or language understanding. For KidsNanny, the full pipeline (Stage~1+2) operates with OCR and 7B LLM reasoning active. Comparison models: ShieldGemma-2~\cite{shieldgemma2} ($\sim$4B parameters) and LlavaGuard~\cite{helff2025llavaguard} ($\sim$7B parameters).

This design ensures each comparison is architecturally aligned and isolates the contribution of Stage~2.

\subsection{Text-Containing Sub-Benchmarks}
\label{sec:text_sub_benchmarks}

To evaluate text-awareness capabilities --- the primary motivation for Stage~2 --- we analyze the 1,054-image test set and identify images containing embedded text. Subset membership was determined by human annotators (team members, blind to model outputs during labeling) who labeled each image based on whether embedded text was present and whether that text constituted the primary safety-relevant signal. This yielded two filtered subsets:

\begin{itemize}[leftmargin=1.5em]
  \item \textbf{Text+Visual} (257 images, 155 unsafe, 102 safe): Images containing both visual content and embedded text.
  \item \textbf{Text-Only} (44 images, 25 unsafe, 19 safe): Images where the safety-relevant signal is primarily or exclusively conveyed through embedded text, with minimal visual explicitness.
\end{itemize}

These subsets are evaluated using Regime~2 models only (KidsNanny Stage~1+2, ShieldGemma-2, LlavaGuard), as vision-only models have no text understanding capability.

\subsection{Evaluation Protocol}
\label{sec:eval_protocol}

All comparison models are evaluated using publicly available pre-trained checkpoints with default inference configurations. No fine-tuning or retraining was performed.

\textit{Note on self-reported vs.\ independently measured results:} ShieldGemma-2's authors report an F1 of 64.2\% on UnsafeBench Sexual in their paper~\cite{shieldgemma2}, while our independent evaluation yields 76.68\%. Differences may arise from evaluation threshold selection, prompt template, or test-set composition. Our evaluation applies a consistent binary classification threshold across all models on the same 1,054-image set.

\textbf{Model identifiers:}

\begin{center}
\small
\begin{tabular}{@{}ll@{}}
\toprule
\textbf{Model} & \textbf{HuggingFace / Source} \\
\midrule
NudeNet & notAI-tech/NudeNet \\
FalconsAI & Falconsai/nsfw\_image\_detection \\
Adam-ViT & AdamCodd/vit-base-nsfw-detector \\
Freepik & Freepik/nsfw\_image\_detector \\
ShieldGemma-2 & google/shieldgemma-2-4b-it \\
LlavaGuard & AIML-TUDA/LlavaGuard-v1.2-7B-OV-hf \\
\bottomrule
\end{tabular}
\end{center}

All inference times are averaged over all 1,054 images on a single NVIDIA RTX 4090 GPU with CUDA acceleration at batch size~1, with 3 GPU warm-up iterations discarded. Metrics: Accuracy, Precision, Recall, and F1 Score~\cite{powers2011evaluation}.

\subsection{Comparison Models}
\label{sec:comparison_models}

\textbf{Vision-only models (Regime~1):}
\begin{itemize}[leftmargin=1.5em]
  \item \textbf{FalconsAI}~\cite{falconsainsfw}: ViT-based binary NSFW classifier, $\sim$85M parameters.
  \item \textbf{Adam-ViT}~\cite{adamcoddvitnsfw}: ViT-B/16 fine-tuned for binary NSFW detection, $\sim$85M parameters.
  \item \textbf{NudeNet}~\cite{nudenet}: Nudity-specific detector using object detection for body part localization.
  \item \textbf{Freepik}~\cite{freepiknsfwdetector}: ViT-based NSFW image classifier.
\end{itemize}

\textbf{Multimodal models (Regime~2):}
\begin{itemize}[leftmargin=1.5em]
  \item \textbf{ShieldGemma-2}~\cite{shieldgemma2}: Gemma~3 4B fine-tuned for multi-category content safety.
  \item \textbf{LlavaGuard v1.2}~\cite{helff2025llavaguard}: LLaVA-OneVision-based 7B VLM for policy-customizable safety assessment.
\end{itemize}
\textit{To the best of our knowledge, ShieldGemma-2 and LlavaGuard v1.2 are the principal open-source VLMs with published safety-specific fine-tuning available at the time of evaluation.}

All models were evaluated under identical conditions: datasets used as-is, same NVIDIA RTX 4090 GPU, same 1,054 images across all regimes.

\subsection{Functional Capability Comparison}
\label{sec:capability_comparison}

Beyond quantitative metrics, content moderation systems differ in the threat vectors they can detect. Table~\ref{tab:capability_comparison} compares functional capabilities across all evaluated models.

\begin{table*}[!t]
\centering
\caption{Functional capability comparison across evaluated models.
$\bullet$~=~evaluated in this paper;
Partial~=~incidental capability, not a dedicated pipeline;
$\times$~=~not supported.}
\label{tab:capability_comparison}
\footnotesize
\setlength{\tabcolsep}{4pt}
\begin{tabular}{@{}lccccccc@{}}
\toprule
\textbf{Capability}
  & \textbf{NudeNet} & \textbf{FalconsAI} & \textbf{Adam-ViT}
  & \textbf{Freepik} & \textbf{ShieldGemma-2} & \textbf{LlavaGuard}
  & \textbf{KidsNanny} \\
\midrule
Image classification
  & $\bullet$ & $\bullet$ & $\bullet$ & $\bullet$
  & $\bullet$ & $\bullet$ & $\bullet$ \\
NSFW object detection
  & $\bullet$* & $\times$ & $\times$ & $\times$
  & $\times$ & $\times$ & $\bullet$ \\
Dedicated OCR pipeline
  & $\times$ & $\times$ & $\times$ & $\times$
  & Partial$^\dagger$ & Partial$^\dagger$ & $\bullet$ \\
Text safety analysis
  & $\times$ & $\times$ & $\times$ & $\times$
  & Partial$^\dagger$ & Partial$^\dagger$ & $\bullet$ \\
Explicit reasoning output
  & $\times$ & $\times$ & $\times$ & $\times$
  & $\times$ & $\bullet$ & $\bullet$ \\
\midrule
Inference time (ms)
  & 35 & 7.3 & 7.2 & 15.5
  & 1{,}136 & 4{,}138 & 11.7\,/\,120$^\ddagger$ \\
\bottomrule
\end{tabular}
\par\smallskip
\footnotesize
* NudeNet performs object detection for nudity-specific body parts only.\\
$^\dagger$ VLMs may read text incidentally via their vision encoder; no dedicated OCR pipeline is employed.\\
$^\ddagger$ 11.7\,ms = Stage~1 only; 120\,ms = full Stage~1+2 pipeline.
\end{table*}

\section{Results}
\label{sec:results}

\subsection{Main Results}
\label{sec:regime1_results}
\label{sec:regime2_results}
\label{sec:cross_regime}

Table~\ref{tab:results_overview} summarises all evaluation results: Regime~1 (vision-only), Regime~2 (multimodal), and Stage~2 contribution across evaluation regimes.

\begin{table*}[!t]
\centering
\caption{Comprehensive evaluation results on UnsafeBench Sexual category (1,054 images). Bold = best in group. $\Delta$ row shows Stage~2 net contribution over Stage~1.}
\label{tab:results_overview}
\small
\resizebox{\textwidth}{!}{%
\begin{tabular}{@{}lccccr@{}}
\toprule
\textbf{Model / Configuration} & \textbf{Accuracy (\%)} & \textbf{F1 (\%)} & \textbf{Precision (\%)} & \textbf{Recall (\%)} & \textbf{Inference (ms)} \\
\midrule
\multicolumn{6}{@{}l}{\textit{Regime~1 --- Vision-Only (Stage~1 active; OCR and LLM disabled)}} \\[1pt]
\midrule
FalconsAI            & 59.01          & 56.28          & 91.15          & 40.70          & 7.3  \\
NudeNet              & 68.03          & 76.01          & 73.96          & 78.18          & 35   \\
Adam-ViT             & 68.98          & 73.90          & 81.23          & 67.79          & 7.2  \\
Freepik              & 77.04          & 81.12          & 86.81          & 76.13          & 15.5 \\
\textbf{KidsNanny Stage~1} & \textbf{80.27} & \textbf{85.39} & 82.05 & \textbf{89.02} & 11.7 \\
\midrule
\multicolumn{6}{@{}l}{\textit{Regime~2 --- Multimodal (full Stage~1+2 pipeline)}} \\[1pt]
\midrule
ShieldGemma-2              & 64.80          & 76.68          & 64.96 & 93.56          & 1136 \\
LlavaGuard                 & 80.36          & 84.56          & 86.17 & 83.02          & 4138 \\
\textbf{KidsNanny (Stage~1+2)} & \textbf{81.40} & \textbf{86.16} & 83.22 & 89.31 & \textbf{120} \\
\midrule
\multicolumn{6}{@{}l}{\textit{Stage~2 contribution (cross-regime comparison, KidsNanny only)}} \\[1pt]
\midrule
KidsNanny Stage~1          & 80.27          & 85.39          & 82.05          & 89.02          & 11.7       \\
KidsNanny (Stage~1+2)      & 81.40          & 86.16          & 83.22          & 89.31          & 120        \\
\textbf{$\Delta$ Stage~2}  & \textbf{+1.13} & \textbf{+0.77} & \textbf{+1.17} & \textbf{+0.29} & $\approx$+108 \\
\bottomrule
\end{tabular}%
}
\end{table*}

\textbf{Regime~1.} KidsNanny Stage~1 achieves the highest accuracy (80.27\%), F1 (85.39\%), and recall (89.02\%) among all vision-only models. Freepik is the closest competitor (77.04\% accuracy, 81.12\% F1). FalconsAI achieves the highest precision (91.15\%) but misses over 59\% of unsafe content. Adam-ViT is the fastest at 7.2\,ms but achieves lower accuracy (68.98\%).

\textbf{Regime~2.} KidsNanny (Stage~1+2) achieves the highest accuracy (81.40\%) and F1 (86.16\%), surpassing LlavaGuard (80.36\%, 84.56\% F1) and ShieldGemma-2 (64.80\%, 76.68\% F1). KidsNanny's full pipeline at 120\,ms is approximately 9$\times$ faster than ShieldGemma-2 (1,136\,ms) and 34$\times$ faster than LlavaGuard (4,138\,ms).

\textbf{Stage~2 contribution.} On the full dataset, Stage~2 adds +1.13~pp accuracy and +1.17~pp precision, reducing false positives from 133 to 123, at $\approx$108\,ms additional latency. Stage~2's impact is far more pronounced on text-containing images (Section~\ref{sec:text_subset}), where it provides the primary detection mechanism.

\subsection{Text-Subset Analysis: Evaluating OCR-Based Detection}
\label{sec:text_subset}

To evaluate Stage~2's text-awareness capability, we benchmark all Regime~2 models on the text-containing subsets from Section~\ref{sec:text_sub_benchmarks}. Table~\ref{tab:text_subsets} reports results.

\begin{table*}[!t]
\centering
\caption{Text-subset evaluation (Regime~2 models only). Top: Text+Visual (257 images, 155 unsafe, 102 safe). Bottom: Text-Only (44 images, 25 unsafe, 19 safe).}
\label{tab:text_subsets}
\small
\resizebox{\textwidth}{!}{%
\begin{tabular}{@{}lccccr@{}}
\toprule
\textbf{Model} & \textbf{Accuracy (\%)} & \textbf{F1 (\%)} & \textbf{Precision (\%)} & \textbf{Recall (\%)} & \textbf{Inference (ms)} \\
\midrule
\multicolumn{6}{@{}l}{\textit{Text+Visual subset (257 images --- visual content with embedded text)}} \\[1pt]
\midrule
ShieldGemma-2 & 54.86 & 51.67 & 72.94 & 40.00 & 1136 \\
LlavaGuard & 76.45 & 79.18 & 84.06 & 74.84 & 4138 \\
\textbf{KidsNanny (Stage~1+2)} & \textbf{81.08} & \textbf{85.11} & 80.46 & \textbf{90.32} & \textbf{115} \\
\midrule
\multicolumn{6}{@{}l}{\textit{Text-Only subset (44 images --- safety depends primarily on embedded text)}} \\[1pt]
\midrule
ShieldGemma-2 & 59.09 & 70.00 & 60.00 & 84.00 & 1136 \\
LlavaGuard & 59.09 & 60.87 & 66.67 & 56.00 & 4138 \\
\textbf{KidsNanny (Stage~1+2)} & \textbf{81.82} & \textbf{86.21} & 75.76 & \textbf{100.00} & \textbf{122} \\
\bottomrule
\end{tabular}%
}
\end{table*}

On the text+visual subset, KidsNanny achieves the highest accuracy (81.08\%), F1 (85.11\%), and recall (90.32\%). ShieldGemma-2's accuracy drops sharply (64.80\%\,$\to$\,54.86\%), confirming that embedded text is particularly challenging for vision-language encoders. On text-only images, KidsNanny's dedicated OCR pipeline achieves 100\% recall (25/25 positives; small sample, interpret cautiously) --- detecting every text-only threat --- while ShieldGemma-2 achieves 84\% recall at substantially lower precision (60\% vs.\ 75.76\%) and approximately 9$\times$ higher latency. LlavaGuard achieves only 56\% recall on text-only threats. KidsNanny's OCR-based approach thus provides the best combination of recall, precision, and efficiency on text-embedded threats.

\subsection{Comparative Visualizations}
\label{sec:visualizations}

\begin{figure*}[!t]
\centering
\phantomsection\label{fig:accuracy_f1}%
\phantomsection\label{fig:precision_recall}%
\phantomsection\label{fig:pareto_frontier}%
\includegraphics[width=0.88\textwidth]{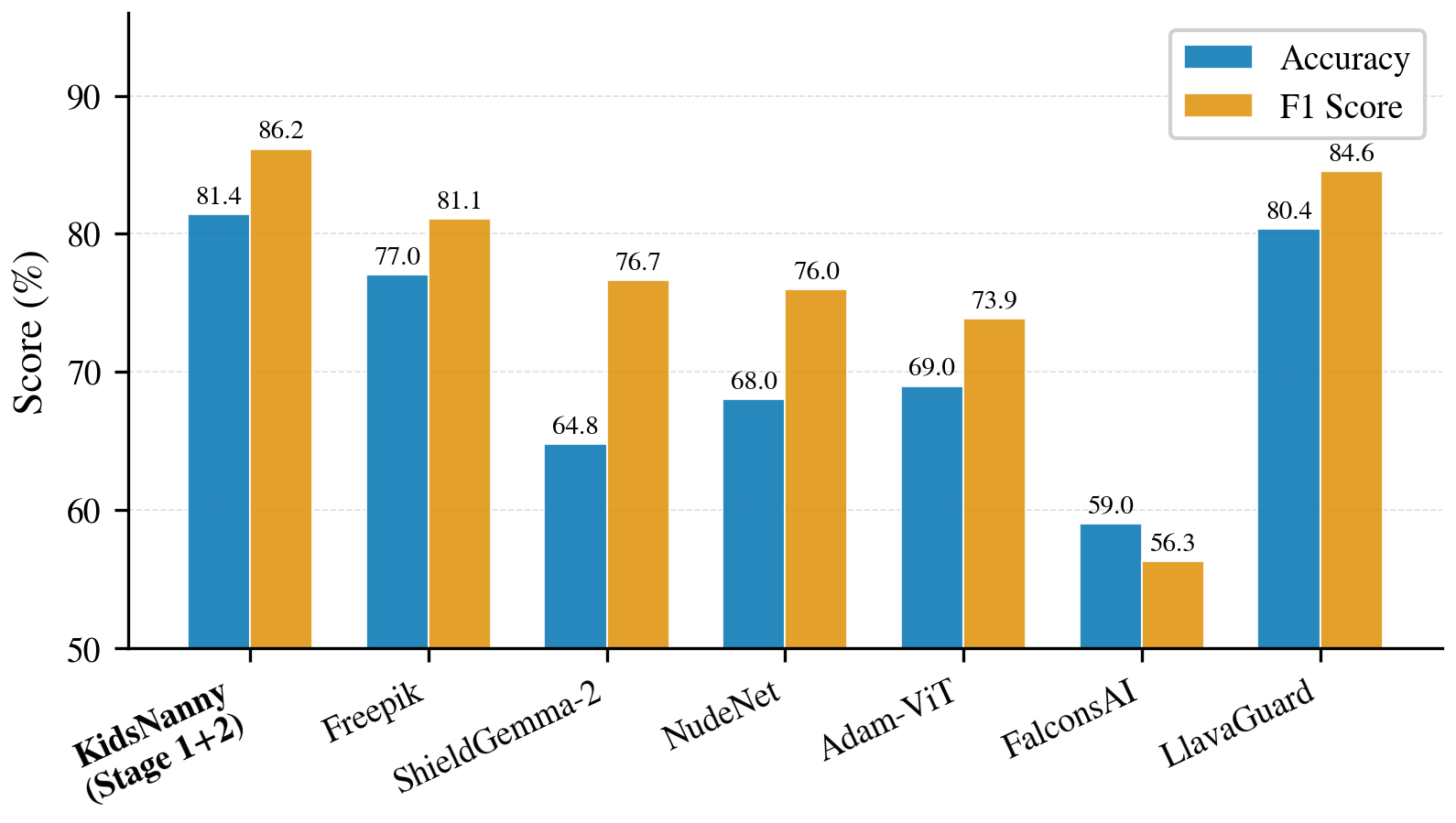}\\[4pt]
\small\textit{(a) Accuracy and F1 Score on UnsafeBench Sexual category (y-axis truncated at 50\%). KidsNanny shows Stage~1+2 (multimodal) results. Models span both evaluation regimes; see Table~\ref{tab:results_overview} for regime-separated comparisons.}

\vspace{8pt}

\begin{minipage}[t]{0.475\textwidth}
  \centering
  \includegraphics[width=\linewidth]{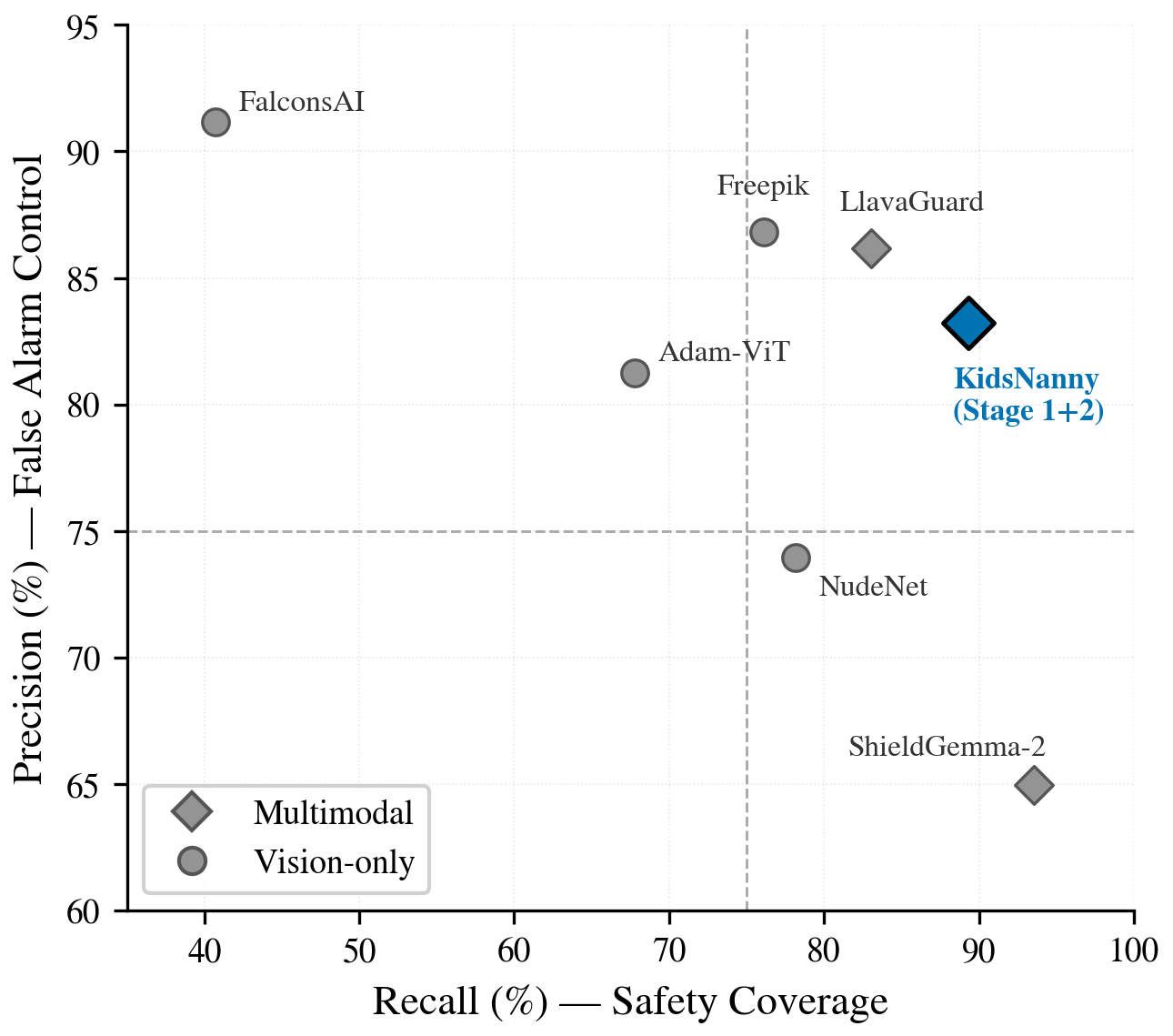}\\[2pt]
  \small\textit{(b) Precision vs.\ Recall trade-off. Dashed lines at 75\% show deployment-viable zone.}
\end{minipage}
\hfill
\begin{minipage}[t]{0.475\textwidth}
  \centering
  \includegraphics[width=\linewidth]{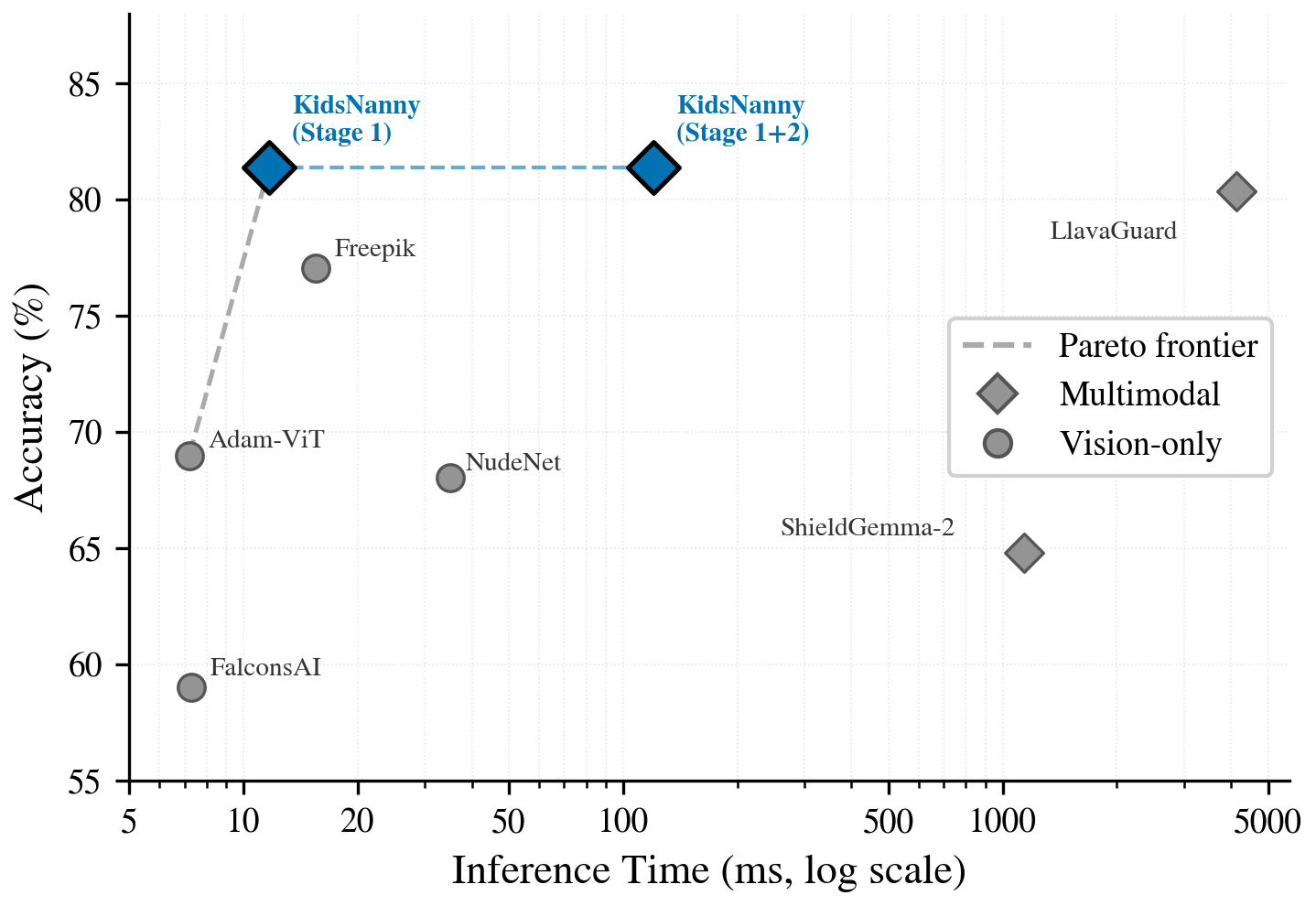}\\[2pt]
  \small\textit{(c) Pareto frontier: Accuracy vs.\ Inference Time (log scale). NVIDIA RTX 4090.}
\end{minipage}

\caption{Model performance visualizations. Diamond markers = multimodal models; circles = vision-only models. (a)~KidsNanny (Stage~1+2) leads all models in accuracy (81.40\%) and F1 (86.16\%). (b)~KidsNanny occupies the balanced deployment-viable zone; ShieldGemma-2 achieves the highest recall but at low precision (64.96\%). (c)~KidsNanny Stage~1 (11.7\,ms) lies on the Pareto frontier; Stage~1+2 (120\,ms) is the overall accuracy leader at 9--34$\times$ lower latency than competing VLMs.}
\label{fig:visualizations}
\end{figure*}

Figure~\ref{fig:visualizations} presents model performance across three complementary dimensions: classification quality, precision--recall trade-off, and latency efficiency.

Figure~\ref{fig:visualizations}(a) shows that KidsNanny (Stage~1+2) achieves the highest accuracy (81.40\%) and F1 score (86.16\%), followed closely by LlavaGuard (80.36\% accuracy, 84.56\% F1). LlavaGuard's inference time of 4,138\,ms --- approximately 34$\times$ that of KidsNanny --- renders it impractical for real-time deployments. Figure~\ref{fig:visualizations}(b) shows that ShieldGemma-2 achieves the highest recall (93.56\%) but its precision of 64.96\% results in substantially more false positives; FalconsAI achieves the highest precision (91.15\%) but misses over 59\% of unsafe content. Figure~\ref{fig:visualizations}(c) confirms that KidsNanny Stage~1 (80.27\%, 11.7\,ms) lies on the Pareto frontier among vision-only models, with Stage~1+2 (81.40\%, 120\,ms) as the overall accuracy leader.

\subsection{Multimodal Model Trade-off Comparison}
\label{sec:competitive}

\begin{table*}[!t]
\centering
\caption{Safety philosophy and trade-off comparison (multimodal regime).}
\label{tab:safety_philosophy}
\small
\begin{tabular}{@{}lp{3.2cm}p{3.2cm}p{2.8cm}@{}}
\toprule
\textbf{Dimension} & \textbf{KidsNanny} & \textbf{ShieldGemma-2} & \textbf{LlavaGuard} \\
\midrule
Primary focus     & Balanced (FN+FP)         & Recall maximisation      & Balanced accuracy \\
Recall (full)     & 89.31\%                  & 93.56\% (highest)        & 83.02\% \\
Recall (text-only)& \textbf{100.00\%}        & 84.00\%                  & 56.00\% \\
Precision         & 83.22\%                  & 64.96\%                  & 86.17\% \\
Inference speed   & 120\,ms                  & 1,136\,ms                & 4,138\,ms \\
Context input     & ViT + Object + OCR       & Vision encoder (\mbox{incidental} text) & Vision encoder \\
\bottomrule
\end{tabular}
\end{table*}

The text-only recall row shows that while ShieldGemma-2 achieves 84\% recall, it does so with substantially lower precision (60\% vs.\ 75.76\% for KidsNanny), resulting in a high false positive rate on safe text-containing images. KidsNanny's dedicated OCR pipeline achieves the highest recall (100\%) while maintaining higher precision than ShieldGemma-2, demonstrating the advantage of explicit text extraction over implicit VLM-based text processing.

\section{Discussion}
\label{sec:discussion}

\subsection{Performance in Context}
\label{sec:performance_context}

KidsNanny achieves the highest accuracy and F1 score in both evaluation regimes. In the vision-only regime, Stage~1 outperforms four dedicated image classifiers (Table~\ref{tab:results_overview}, Regime~1). In the multimodal regime, the full pipeline surpasses both ShieldGemma-2 and LlavaGuard while operating at 9--34$\times$ lower inference latency (Table~\ref{tab:results_overview}, Regime~2).

The two-regime design reveals that Stage~2's contribution on the full dataset is primarily precision refinement --- reducing false positives by 7.5\% (10 of 133) at negligible recall cost. However, the text-subset analysis reveals a far more significant contribution: on text-only images, Stage~2 provides the primary detection mechanism, achieving 100\% recall where VLMs achieve 56--84\%.

\subsection{Architectural Contribution}
\label{sec:arch_contribution}

KidsNanny's key architectural contribution is routing only text and object labels --- not raw pixels --- to the reasoning LLM. This achieves two advantages simultaneously: (1)~120\,ms full-pipeline inference versus 1,136--4,138\,ms for VLMs, and (2)~superior text-based threat detection, because dedicated OCR may extract safety-relevant text more reliably than VLM vision encoders on this benchmark, consistent with ShieldGemma-2's authors' own acknowledgement~\cite{shieldgemma2} of text-visual interplay limitations. Qu et al.~\cite{qu2025modalitygap} document a concrete mechanism underlying this gap: VLMs attend to background scene elements --- objects, people, environments --- rather than text overlaid on images (e.g., focusing on humans and laptops in cyberbullying content while ignoring the offensive text overlay), a behavioural bias attributable in part to unsafe images constituting only 3--6\% of public vision pretraining corpora.

The text-only subset results (Table~\ref{tab:text_subsets}, bottom) provide strong evidence for this architectural choice: while ShieldGemma-2 achieves 84\% recall on text-only threats, it suffers from low precision (60\%) due to high false positives, and requires 1,136\,ms inference. KidsNanny's dedicated OCR pipeline achieves 100\% recall with 75.76\% precision at 122\,ms --- suggesting that, on this benchmark, explicit text extraction achieves a better recall-precision balance than the VLM-based comparators evaluated here, though statistical significance has not been established and generalisation to other datasets remains to be confirmed.

\subsection{Stage~2 Evaluation and Remaining Gaps}
\label{sec:discussion_stage2}

UnsafeBench includes images with embedded text, enabling us to filter text-containing subsets and provide the first quantitative evidence of KidsNanny's OCR-based detection advantage. By separately benchmarking the text+visual (257 images) and text-only (44 images) subsets, we isolate Stage~2's contribution to safety classification.

However, the text-only subset remains small (44 images), limiting statistical power. Larger, purpose-built text-in-image safety benchmarks would strengthen future evaluations of multimodal moderation systems (Section~\ref{sec:future_work}).

\section{Limitations}
\label{sec:limitations}
\begin{itemize}[leftmargin=1.5em]
  \item \textbf{Self-evaluation bias.} This paper was authored, evaluated, and reported entirely by the KidsNanny development team. The benchmark design, subset selection, and result interpretation were all performed without external oversight. This constitutes a direct conflict of interest: the team evaluating the system is the same team that built it. Results should be treated as a first-party technical report pending independent external validation.
  \item \textbf{Metric-specific trade-offs.} KidsNanny does not lead on every metric in isolation. ShieldGemma-2 surpasses it in recall on the full dataset; FalconsAI in precision; Adam-ViT in raw inference speed.
  \item \textbf{Focused evaluation scope.} This study evaluates KidsNanny on the Sexual content category of UnsafeBench, which represents the primary threat category in child safety contexts. Broader evaluation across additional categories such as violence, substance-related content, and others is a natural direction for future work.
  \item \textbf{Small text-only subset.} The text-only sub-benchmark contains 44 images (25 unsafe, 19 safe), limiting statistical confidence. These results should be confirmed on larger, purpose-built text-in-image safety datasets.
  \item \textbf{Proprietary model and limited reproducibility.} KidsNanny is a proprietary system. Exact backbone architectures, parameter counts, and training hyperparameters are withheld, which precludes independent assessment of whether performance stems from architectural novelty or from scale and training data. The reproducible scientific contributions are the two-regime evaluation design and the text-subset benchmarking methodology; the model itself cannot be independently replicated or audited.
  \item \textbf{Single hardware configuration.} All benchmarks were evaluated on an NVIDIA RTX 4090 GPU. Performance may differ on other hardware.
  \item \textbf{Statistical significance.} Pairwise significance tests were not performed; whether observed metric differences are statistically significant remains to be established in future work.
\end{itemize}

\section{Conclusion}
\label{sec:conclusion}

Through systematic benchmarking under two architecturally aligned evaluation regimes, this paper demonstrates that KidsNanny achieves the highest accuracy and F1 score in both vision-only and multimodal evaluations on the UnsafeBench Sexual category. KidsNanny Stage~1 achieves 80.27\% accuracy at 11.7\,ms; adding Stage~2 yields 81.40\% accuracy at 120\,ms --- a modest +1.13 percentage point gain on the full 1,054-image dataset, with Stage~2's primary contribution being precision refinement (reducing false positives by 10 images) rather than a large aggregate accuracy shift.

Stage~2's more substantial contribution emerges on text-containing content: text-subset analysis reveals that KidsNanny's dedicated OCR pipeline achieves 100\% recall (25/25 positives; n\,=\,44) with 75.76\% precision on text-only safety threats --- results that should be interpreted cautiously given the small sample size, but which are consistent with the architectural expectation that explicit OCR outperforms implicit pixel-level text understanding. ShieldGemma-2 reaches 84\% recall on the same subset but with only 60\% precision and approximately 9$\times$ higher latency; LlavaGuard achieves 56\% recall.

These results suggest that the two-stage design is most beneficial for content where safety depends on embedded text, and that its full-dataset advantage over VLM-based approaches lies primarily in latency efficiency rather than raw accuracy gains.

\section{Future Work}
\label{sec:future_work}
\begin{itemize}[leftmargin=1.5em]
  \item \textbf{Multi-category benchmarking.} Extend evaluation to violence, drugs, smoking, self-harm, and other risk categories.
  \item \textbf{Dedicated text-in-image safety benchmarks.} Develop purpose-built test sets for text-embedded threats at scale, addressing the 44-image limitation of the current text-only sub-benchmark.
  \item \textbf{Statistical significance testing.} Perform McNemar's test and bootstrap confidence intervals across all comparisons.
  \item \textbf{Independent third-party validation.} Pursue external evaluation through academic collaborations.
  \item \textbf{Cross-hardware portability.} Evaluate across CPU-only, mobile edge, and cloud GPU configurations.
  \item \textbf{Adversarial robustness.} Assess resilience against perturbation-based evasion, steganographic embedding, and prompt injection.
  \item \textbf{Bias and fairness analysis.} Evaluate detection across demographic groups and cultural contexts, aligned with EU AI Act~\cite{euaiact2024} requirements.
\end{itemize}

\section*{Acknowledgments}
\addcontentsline{toc}{section}{Acknowledgments}

We thank Yiting Qu and the UnsafeBench team at CISPA Helmholtz Center for Information Security for granting permission to use the UnsafeBench dataset. We also thank the Visual Geometry Group at the University of Oxford for making the PASS dataset~\cite{passvggdataset} publicly available. We acknowledge the teams behind ShieldGemma~2, LlavaGuard, NudeNet, Freepik, FalconsAI, and Adam-ViT for making their models publicly available for evaluation.

\medskip
\noindent\textit{Permission:} UnsafeBench benchmarking use was conducted with written permission from the dataset authors, obtained February 2026.

\bibliographystyle{plain}
\bibliography{references}

\end{document}